# Boosting Unsupervised Domain Adaptation with Soft Pseudo-label and Curriculum Learning

*ZHANG Shengjia* (张晟嘉)[1, 2], LIN Tiancheng (林天成)[1, 2], *XU Yi*[*] (徐奕)[1, 21]

(1. School of Electronic Information and Electrical Engineering, Shanghai Jiao Tong University, Shanghai 200240, China, 2. Shanghai Key Laboratory of Digital Media Processing and Transmission Shanghai, China)

**Abstract:** By leveraging data from a fully labeled source domain, unsupervised domain adaptation (UDA) improves classification performance on an unlabeled target domain through explicit discrepancy minimization of data distribution or adversarial learning. As an enhancement, category alignment is involved during adaptation to reinforce target feature discrimination by utilizing model prediction. However, there remain unexplored problems about pseudo-label inaccuracy incurred by wrong category predictions on target domain, and distribution deviation caused by overfitting on source domain. In this paper, we propose a model-agnostic two-stage learning framework, which greatly reduces flawed model predictions using soft pseudo-label strategy and avoids overfitting on source domain with a curriculum learning strategy. Theoretically, it successfully decreases the combined risk in the upper bound of expected error on the target domain. At the first stage, we train a model with distribution alignment-based UDA method to obtain soft semantic label on target domain with rather high confidence. To avoid overfitting on source domain, at the second stage, we propose a curriculum learning strategy to adaptively control the weighting between losses from the two domains so that the focus of the training stage is gradually shifted from source distribution to target distribution with prediction confidence boosted on the target domain. Extensive experiments on two well-known benchmark datasets validate the universal effectiveness of our proposed framework on promoting the performance of the top-ranked UDA algorithms and demonstrate its consistent superior performance.

* **E-mail:** xuyi@sjtu.edu.cn

## 0 Introduction

When the training dataset and testing dataset share a similar data distribution, deep Convolutional Neural Networks (CNNs) can achieve expert-level performance in image classification tasks. However, applying a trained model directly to an unseen testing dataset usually suffers from obvious performance degradation. That phenomenon is rooted in the data distribution difference between source domain (training dataset) and target domain (testing dataset), known as *domain shift*[1-2]. Although collecting and annotating abundant data on the target domain is a possible solution, that is time-consuming and even needs professional annotation. In order to tackle that problem, recently a large number of unsupervised domain adaptation (UDA) algorithms have been proposed[3-13]. According to UDA theory[14-16], the learning bound consists of source risk, marginal distribution discrepancy and combined risk (details in Section 2.2). Based on the theory, source risk is easily controlled with ground truth on source domain. Combined risk, representing *adaptability* between source and target domain, is a negligibly small constant. Therefore, mainstream UDA approaches focus on minimizing marginal distribution discrepancy so that the model trained on a fully labeled source domain can be applied to an unlabeled target domain.

With the intention of aligning the distribution between source domain and target domain, mainstream UDA algorithms can be generally divided into two kinds: direct feature distribution alignment and adversarial domain-invariant feature learning. Methods of the former category diminish discrepancy between feature distributions of source domain and target domain with explicit metrics[17-22], *e.g.*, correlation alignment[23-24] or moment matching[21-22]. The latter category of algorithms is inspired by Generative Adversarial Network (GAN)[25] and enforces feature distribution alignment by making image representations indistinguishable to a domain discriminator[4,7,26-32]. While globally aligning the source and target distributions is beneficial for domain adaptation, the relationship between two corresponding categories in source and target domain is not taken into account. Consequently, all the data from both domains will be mixed up regardless of their classes during adaptation, and discriminative structures along with category-related representations would be destroyed. In order to maintain the domain structure and preserve the fine-grained category information in the adaptation process, category alignment has been introduced lately as an additional restriction to global distribution alignment. As a complement to previous distribution alignment methods (both explicit feature distribution alignment and adversarial domain-invariant feature learning), model predictions on target domain are utilized as a supplementary condition for category alignment[3,5-6,20,33-40]. Nevertheless, as a gap exists between the model prediction and ground truth, incorrect hard pseudo-labels on target domain can mislead the adaptation training. Moreover, considering ground truth (one-hot) supervision on source domain, it is a prevalent problem that model with weak supervision on target domain overfits to source distribution[41]. In that situation, the combined risk is no longer negligible and it degrades the adaptation performance[12].

In this work, we aim to tackle these two challenging problems of UDA, *i.e.*, false label predictions on target domain misguiding the adaptation, and overfitting to source distribution owing to ground truth dominance. That is a paradox which cannot be solved in one-stage training because current UDA algorithms predict pseudo-labels on target domain by learning the source distribution. Therefore, we propose a model-agnostic two-stage learning framework with Soft Pseudo-label and Curriculum Learning (SPCL) as illustrated in Fig.1.

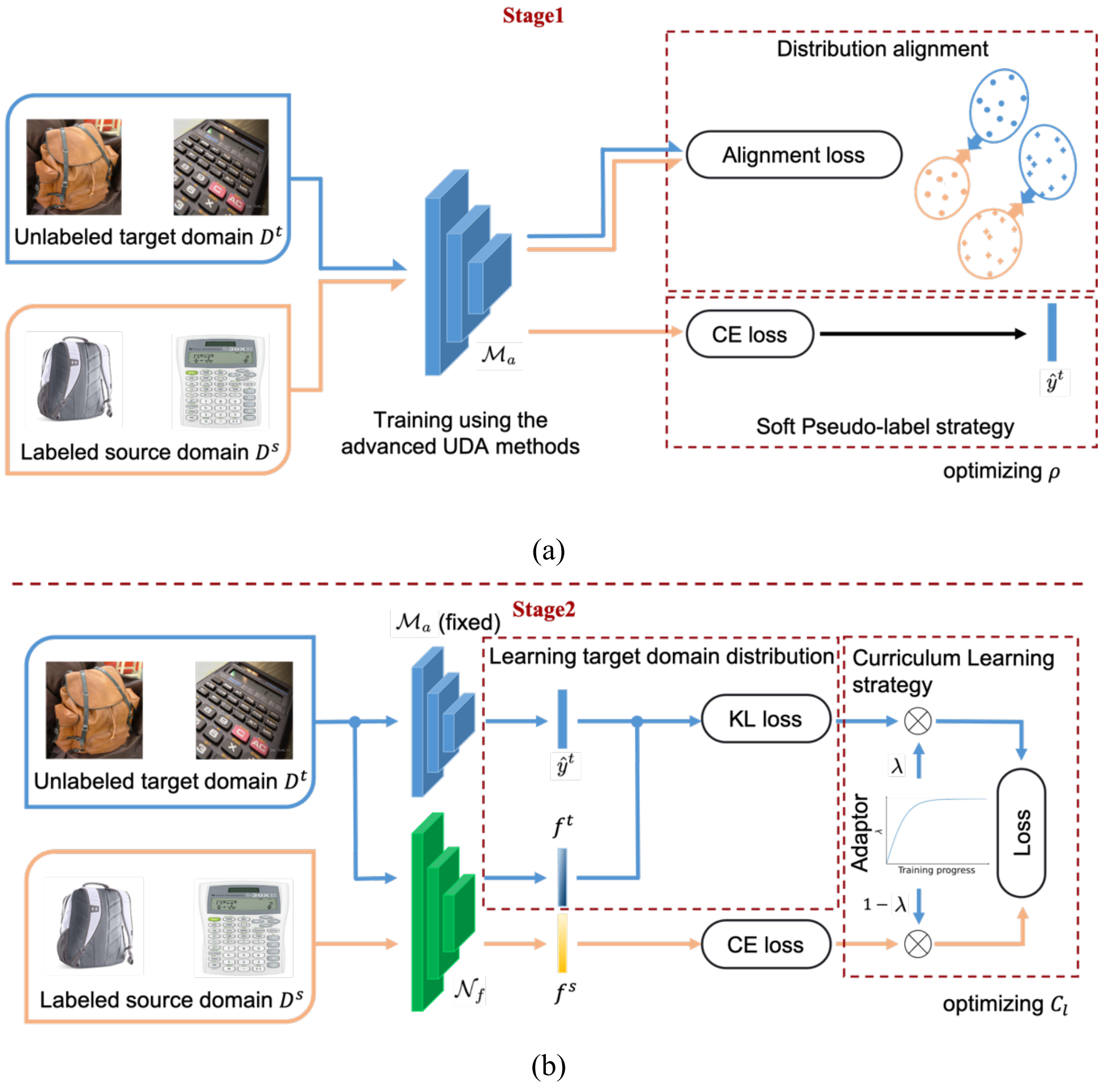

Fig. 1 Framework of our SPCL.

Our contributions are: 1) designing a soft pseudo-label strategy to obtain soft labels on target domain with rather high confidence via a trained UDA model that aligns source data distribution and target data distribution at the first stage; 2) proposing a curriculum learning strategy at the second stage to further improve classification performance; 3) SPCL is universally effective for various UDA algorithms and achieves SOTA classification accuracy.

It should be noted that, SPCL adopts soft pseudo-label strategy instead of using one-hot pseudo-labels. Compared with hard pseudo-labels, soft labels alleviate the misguidance of inaccurate model predictions on the target domain. More importantly, soft labels convey and emphasize task-related semantic information[42-44]. Furthermore, we design a curriculum learning strategy to avoid overfitting the source distribution at the second stage, where we gradually shift training focus from source distribution to target distribution. Specifically, an adaptor controls the loss weighting between source domain and target domain with a gradual update strategy. At the beginning of the second stage, source domain will dominate the training so as to build a preliminary network with ground truth information. Then, as the training proceeds, the network learns rather reliable soft pseudo-labels and accordingly the weighting of target domain is increased, enforcing the network to pay increasing attention to target domain. Significantly, we minimize the combined risk using our soft pseudo-label strategy and curriculum learning strategy, successfully decreasing the upper bound of expected error on target domain.

The remainder of this paper is structured as follows. In Section 1, we introduce related works on distribution alignment-based UDA and the intention of our proposed SPCL. In Section 2, we provide the theoretical analysis to prove that SPCL successfully decreases the upper bound of expected error on target domain through reducing the combined risk and thus promisingly boosts adaptation performance. Also, we explain the details of our soft pseudo-label strategy and curriculum learning strategy. In Section 3, extensive experiments are conducted on two well-known benchmark classification datasets (Office-31[45] and Office-Home[46]). Our proposed SPCL is applied to various *state-of-the-art* (SOTA) UDA models (DAN[18], DSAN[3], CAN[6], GVB-GD[4]) with consistent performance improvement.

## 1 Related Work

In this section, we will introduce the related work in three aspects, including distribution alignment for domain adaptation, category prediction to enhance adaptation and curriculum learning to avoid overfitting.

## 1.1 Distribution Alignment for Domain Adaptation

UDA intends to transfer information from a fully annotated source domain to an unlabeled target domain. Recent years have witnessed advances by embedding domain adaptation modules in the pipeline of deep feature learning to extract transferable domain-invariant representations[7-8,18,24,27,33,47]. In the literature, distribution-aligning domain adaptation approaches can be categorized into two common practices, *i.e.*, explicit statistics-based distribution alignment and adversarial learning for distribution alignment.

The former approach aligns feature distribution by minimizing explicit discrepancy metrics. Long *et al.* propose DAN[18] and JAN[48] to minimize Maximum Mean Discrepancy (MMD)[49] and joint MMD distance. Peng *et al.*, Zellinger *et al.* and Li *et al.* advocate moment matching[21-22,50] as distribution similarity metric. Sun *et al.* apply correlation alignment[1,24] to diminish domain divergence with second-order statistics. Those distance-based regularizers are introduced at higher layers of the network between the two representations from source domain and target domain. The latter approach borrows the idea from Generative Adversarial Networks (GANs)[25]. Adversarial learning has gained research focus in domain adaptation[4,7-8,32,34,51-57] to decrease the domain discrepancy or align feature distribution in different domains. Ganin *et al.* propose a Gradient Reversal Layer (GRL) in Domain Adversarial Neural Network (DANN)[27] to confuse the domain classifier, which is designed to distinguish the source domain from the target domain. Other works such as ADDA[54], MCD[56], CyCADA[58] and SBADA[59] are likewise constructing adversarial learning networks to align feature distribution. Gradually Vanishing Bridge (GVB-GD)[4] cooperates with adversarial domain adaptation approaches. GVB-GD enhances the representation ability of the discriminator and suppresses residual domain-specific information in domain-invariant representations on the generator. Another research focus is concentrating on applying GAN as an image synthesizer on target domain[30-31,60-61]. Our framework SPCL is most related to UDA methods of feature distribution aligning. Given an arbitrary distribution aligned model, SPCL can utilize the predictions (soft pseudo-labels) of this model and improve the prediction accuracy in the target domain through curriculum learning at the second stage. It is expected to be universally effective for enhancing the prediction of distribution aligned UDA model.

## 1.2 Enhancement of Category Prediction

As an enhancement to distribution alignment approaches for domain adaptation, recent researches have concentrated on narrowing the divergence between the corresponding categories (subdomains) of source domain and the target domain, contributing to more accurate and fine-grained distribution alignment and preserving information of domain structure and category-related representations[3,7,33,36,62]. Long *et al.* propose Conditional Domain Adversarial Network with Entropy (CDAN+E)[7], conditioning the adversarial adaptation learning on discriminative information involved in classifier predictions. Pei *et al.* utilize Multi-Adversarial Domain Adaptation (MADA)[33] to capture multimode structures in order to align different data distributions accurately with multiple-domain discriminators. Xie *et al.* explore semantic features for the unlabeled target domain with Moving Semantic Transfer Network (MSTN)[36], correlating labeled source centroids with pseudo-labeled target centroids. Kumar *et al.* propose Co-DA[62] to build multiple feature spaces. Co-DA aligns feature distributions of source domain and target domain in each of those feature spaces while constraining that all alignments yield consistent category predictions on unlabeled target data. Besides those adversarial distribution-aligning methods, Zhu *et al.* present a non-adversarial method called Deep Subdomain Adaptation Network (DSAN) to measure the Local Maximum Mean Discrepancy (LMMD) between relevant subdomains on source domain and target domain, which can consider the weight of different samples[3]. However, previous pseudo-label assisted distribution adaptation algorithms make use of hard model predictions during training iterations. Notably, there is a gap existing between the ground truth and the pseudo-labels estimated by the training model on target domain. Therefore, those algorithms are prone to inaccurate model predictions and suffer from the misguidance. As a result, combined risk degrades the whole adaptation performance. Instead of using hard pseudo-labels, SPCL utilizes soft pseudo-labels from an adapted (distribution aligned) model instead, boosting prediction confidence and transferring task-related semantic information.

## 1.3 Curriculum Learning

Curriculum learning is a self-paced strategy that follows an "easy-to-hard" scheme. In the beginning, network is learning information or distribution with rather high confidence and then it moves onto difficult data. In the context of domain adaptation, Zhang *et al.* propose a curriculum adaptation approach, which infers target domain properties based on learning distributions on source domain for semantic segmentation tasks[63]. Zou *et al.* alleviate dominance of large classes on pseudo-label generation and refine generated labels with spatial priors by Class-Balanced Self-Training (CBST)[39]. Jaehoon *et al.* propose Pseudo-labeling Curriculum for unsupervised Domain Adaptation (PCDA)[64] that trains network on samples with high-density values at early stages and on those with low-density values later. Xiao *et al.* balance domain alignment learning and class discrimination learning with Dynamic Weighted Learning (DWL)[9]. Different from aforementioned curriculum learning strategies that balance data weighting in either source domain or target domain, SPCL focuses on regularizing the weighting of source domain and target domain. SPCL successfully alleviates overfitting on the source domain and decreases the upper bound of expected error on the target domain.

## 2 Soft Pseudo-label and Curriculum Learning

In this section, we explore the details of our proposed two-stage framework called SPCL. The overall procedure is depicted in Fig.1. At the first stage, using all labeled source domain data and unlabeled target domain data, $\mathcal{M}_a$ aligns the distribution of source domain and target domain with advanced UDA methods (either statistics-based methods or adversarial learning algorithms), and learns source domain distribution by using labeled source domain data. We adopt Soft Pseudo-label strategy to greatly reduce the inaccurate pseudo-label ratio $\rho$ by utilizing rather accurate soft pseudo-label $\hat{y}^t$ from $\mathcal{M}_a$ on the target domain. At the second stage, we train the network $\mathcal{N}_f$ by learning target domain distribution from $\hat{y}^t$ and learning source domain distribution from labeled source domain data. We adopt Curriculum Learning strategy to reduce the combined risk $C_l$ by using the regularization factor $\lambda$ to solve the overfitting problem on source distribution. The explanation of $\rho$ and $C_l$ will be provided in section 2.2.

### 2.1 Problem Formulation

In the context of UDA, data from *source domain* are fully annotated while data from *target domain* are all unlabeled. Generally, there exists a phenomenon that feature distributions of the two domains are different. By mitigating the distribution discrepancy of source and target data, the objective of image classification task in UDA is to improve performance of the classifier on target domain.

Formally, data from source domain are denoted by $D^s = \left\{\left(\boldsymbol{x}_i^s, y_i^s\right)|_{i=1}^{N^s}\right\}$, where $\boldsymbol{x}_i^s$ is the $i$-th image data from source domain and $y_i^s$ is its corresponding one-hot category label. $N^s$ represents the size of the source dataset. Data from target domain, similarly, are denoted by $D^t = \left\{\boldsymbol{x}_i^t\,|_{i=1}^{N^t}\right\}$, where

$\boldsymbol{x}_i^t$ is the $i$-th image data from target domain and $N^t$ is the number of target data.

UDA algorithms train a model on $D^s \cup D^t$ and aim to obtain high classification accuracy on testing dataset (target domain). Mainstream UDA methods align feature distributions of source and target domain with either explicit statistics-based metric or adversarial learning networks. However, although the fine-grained category alignment has been appended as an enhancement, two challenges are still unexplored for one-stage UDA methods. One is adaptation misguidance due to incorrect model predictions on the target domain. The other one is the overfitting problem on source distribution rooted in ground truth dominance. Both challenges cause an increase in combined risk so that the adaptation performance deteriorates. Therefore, we proposed a model-agnostic two-stage framework SPCL with our soft label strategy and curriculum learning strategy specifically tailored to address those two problems.

## 2.2 Theoretical Insight

Now we analyze the optimization of our proposed two-stage framework SPCL on combined risk, utilizing the theory of domain adaptation[14,35].

*Theorem 1:* With $\mathcal{H}$ being the hypothesis space and given source domain $D^s$ and target domain $D^t$, we have

$$\forall h \in \mathcal{H}, R_{D^t}(h) \le R_{D^s}(h) + \frac{1}{2} d_{\mathcal{H}\Delta\mathcal{H}}(D^s, D^t) + C, \tag{1}$$

where $R_{D^s}(h)$ and $R_{D^t}(h)$ denote expected error on source domain and target domain, respectively. $R_{D^s}(h)$ can be minimized by ground truth on source domain. $d_{\mathcal{H}\Delta\mathcal{H}}(D^s, D^t)$ is the data distribution divergence between source domain $D^s$ and target domain $D^t$. Current UDA algorithms mainly focus on minimizing $d_{\mathcal{H}\Delta\mathcal{H}}(D^s, D^t)$ through explicit statistics-based metrics such as MMD[49], CMD[22] and Deep CORAL[24], and adversarial learning approaches like DANN[27], ADDA[54] and CyCADA[58]. $C$ is the combined risk, which is defined as

$$C = \min_{h \in \mathcal{H}} R_{D^s}\left(h, f_{D^s}\right) + R_{D^t}\left(h, f_{D^t}\right). \tag{2}$$

$f_{D^s}$ and $f_{D^t}$ are true labeling functions for source domain and target domain. Without ground truth function $f_{D^t}$, $C$ cannot be directly measured. Therefore, pseudo-labels are introduced to estimate the combined risk. We denote pseudo-labeled target domain by $D^{t_l} = \left\{(\boldsymbol{x}_i^t, \hat{y}_i^t)\,|_{i=1}^{N^t}\right\}$ and the inaccurate pseudo-label ratio by $\rho$. Note that $\hat{y}_i^t$ is a probability prediction (soft prediction). For convenience we omit ground truth function $f$ and then we have the estimation[35]

$$\begin{aligned} R_{D^s}(h) + R_{D^t}(h) &= R_{D^s}(h) + R_{D^t}(h) - R_{D^{t_l}}(h) + R_{D^{t_l}}(h) \\ &\le R_{D^s}(h) + R_{D^{t_l}}(h) + \left|R_{D^t}(h) - R_{D^{t_l}}(h)\right| \\ &\le R_{D^s}(h) + R_{D^{t_l}}(h) + \rho. \end{aligned} \tag{3}$$

We denote $\min_{h \in \mathcal{H}} R_{D^s}(h) + R_{D^{t_l}}(h)$ by $C_l$, then

$$C \le C_l + \rho, \tag{4}$$

which means that the combined risk of source domain and target domain can be estimated by the summation of combined risk $C_l$ on source domain and pseudo-labeled target domain, and pseudo-label inaccuracy $\rho$. Our proposed framework SPCL utilizes soft pseudo-labels to reduce $\rho$ and adaptively focus on target distribution learning to minimize $C_l$. Consequently, SPCL decreases the upper bound of expected error on target domain by adopting soft pseudo-label strategy and curriculum learning strategy. Our proposed two-stage framework is model-agnostic and expected to be universally effective.

## 2.3 Soft Pseudo-label Strategy

Current UDA algorithms utilize inaccurate hard pseudo-labels during the category alignment for adaptation[4,6], which causes misguidance in domain adaptation. To reduce inaccurate label ratio $\rho$ on target domain, we adopt a soft pseudo-label strategy. Compared with hard pseudo-labels, soft labels from adapted model contain task-relevant semantic information[43] and regularize label smoothing[42,44].

In our proposed framework SPCL, we firstly train a model $\mathcal{M}_a$ with advanced UDA algorithm (DAN[18], DSAN[3], GVB-GD[4] and CAN[6]) at the first stage:

$$\mathcal{L}_{stage1} = \mathcal{L}_{align}\left(D^s, D^t\right) + \mathcal{L}_{CE}\left(\mathcal{M}_a(\boldsymbol{x}^s;\theta_a), y^s\right), \tag{5}$$

where $\mathcal{L}_{align}$ denotes the distribution alignment loss and $\mathcal{L}_{CE}$ (Cross Entropy loss) is the classification loss on source domain; $D^s$, $D^t$ are labeled source domain data and unlabeled target domain data, respectively; $\theta_a$ represents parameters of $\mathcal{M}_a$. For convenience, data index $i$ is omitted. Then soft pseudo-labels $\hat{y}^t$ on target domain are provided by $\mathcal{M}_a$.

$$\hat{y}^t = \mathcal{M}_a(\boldsymbol{x}^t;\theta_a). \tag{6}$$

At the second stage, we transfer target domain knowledge from $\mathcal{M}_a$ to the training network $\mathcal{N}_f$ by minimizing the Kullback–Leibler (KL) divergence between the network prediction on target data with corresponding soft pseudo-label

$$\mathcal{L}_{target} = \mathcal{L}_{KL}\left(\mathcal{N}_f(\boldsymbol{x}^t;\theta_f), \hat{y}^t;\tau\right), \tag{7}$$

where $\tau$ is the hyper-parameter *temperature*. Similarly, $\theta_f$ denotes parameters of $\mathcal{N}_f$. It is noted that the proposed soft pseudo-label strategy alleviates the misguidance of inaccurate one-hot labels and conveys task-related information to the training network. In Section 4.1, we validate the high prediction accuracy of adapted model and compare soft pseudo-label strategy with hard label strategy by experiments and analysis.

## 2.4 Curriculum Learning with Adaptively Weighting Strategy

In order to diminish combined risk $C_l$, we adopt curriculum learning with a gradual weighting strategy. To learn the source distribution, we use

$$\mathcal{L}_{source} = \mathcal{L}_{CE}\left(\mathcal{N}_f(\boldsymbol{x}^s;\theta_f), y^s\right). \tag{8}$$

Because of the one-hot ground truth $y^s$, the training network tends to overfit to the source distribution[41]. To tackle that problem, we design an adaptive weighting strategy to gradually shift the learning focus from source distribution to target distribution. At the second stage, the losses of learning target distribution and source distribution will be regularized by factor $\lambda$, where $\lambda$ is adaptively generated according to the training epoch. Specifically, $T$ represents current epoch and $T_{\max}$ denotes total training epochs. Then $\lambda$ is generated by

$$\lambda = \frac{2}{1+e^{-10\times\frac{T}{T_{max}}}} - 1. \tag{9}$$

The value of $\lambda$ gradually increases as the second stage training proceeds. The adaptive weighting strategy is advanced with two motivations. Firstly, learning source distribution with ground truth label offers a solid objective for the network to converge towards at early stages. Secondly, to enhance classification accuracy on target domain, the network avoids overfitting to the source distribution by predominantly learning rather reliable soft labels on the target distribution. In Section 4.2, we provide qualitative results of different $\lambda$ generation mechanisms.

Overall, the two-stage training of our proposed SPCL can be described as

$$\begin{cases} \mathcal{L}_{stage1} = \mathcal{L}_{align}\left(D^s, D^t\right) + \mathcal{L}_{CE}\left(\mathcal{M}_a(\boldsymbol{x}^s;\theta_a), y^s\right), \\ \mathcal{L}_{stage2} = \alpha\lambda\mathcal{L}_{KL}\left(\mathcal{N}_f(\boldsymbol{x}^t;\theta_f), \hat{y}^t;\tau\right) + (1-\lambda)\mathcal{L}_{CE}\left(\mathcal{N}_f(\boldsymbol{x}^s;\theta_f), y^s\right), \end{cases} \tag{10}$$

where $\alpha$ is a constant to normalize the value range of two losses. With model $\mathcal{M}_a$ trained based on Eqn.(5), the second stage training is summarized in Algorithm 1.

**Algorithm 1** Second stage training of SPCL

**Input:** Labeled source domain data $D^s = \{(\mathbf{x}^s, y^s)\}$, unlabeled target domain data $D^t = \{\mathbf{x}^t\}$, fixed model $\mathcal{M}_a$, total training epochs $T_{max}$, hyper-parameter $\tau$, normalization constant $\alpha$.

1: Initialize $\mathcal{N}_f$ with ImageNet pretrained parameters;
2: Initialize $T = 0$;
3: **while** $T < T_{max}$ **do**
4: $T \leftarrow T + 1$;
5: Get soft pseudo-label $\hat{y}^t$ using Equation (6);
6: Calculate $\lambda$ using Equation (9);
7: Update $\mathcal{N}_f$ by minimizing $\mathcal{L}_{stage2}$ in Equation (10).
8: **end while**

**Output:** $\mathcal{N}_f$

# 3 Experiments

In this section, we evaluate SPCL against competitive advanced UDA algorithms on two challenging standard benchmarks.

## 3.1 Datasets and Settings

**Office-31**[45] is a standard benchmark dataset for visual domain adaptation tasks. It consists of 4,110 images in 31 categories. Three distinct domains are contained: **A**mazon (2,817 images downloaded from amazon.com), **W**ebcam (795 images taken by web camera) and **D**SLR (498 images taken by digital SLR camera). For a fair evaluation, we conduct adaptation on all six transfer tasks: A→W, D→W, W→D, A→D, D→A, W→A.

**Office-Home**[46] is a challenging dataset for visual domain adaptation. It contains 15,848 images in 65 common object classes in home and office scenarios. Office-Home is much larger than Office-31 and has four significantly different domains: **A**rtistic images, **C**lip art, **P**roduct images and **R**eal-world images. Similarly, we construct totally 12 adaptation tasks using all domains.

***Baselines*** We conduct experiments with various SOTA UDA methods. Consequently, for Office-31 we compare our proposed SPCL with ResNet[65], DDC[19], DAN[18], Deep CORAL[24], DANN[27], ADDA[54], JAN[48], MADA[33], MCD[56], GTA[52], DWL[9], iCAN[5], CDAN+E (CDAN with Entropy conditioning)[7], MDD[8], ILA-DA[11], E-MixNet[12], TSA (with batch spectral penalization[66])[13], DSAN[3], CAN (Contrastive Adaptation Network)[6], GVB-GD[4] and MetaAlign (applied to GVB-GD)[10]. For Office-Home, we compare SPCL with ResNet, DAN, DANN, JAN, MCD, CDAN+E, MDD, E-MixNet, TSA, MetaAlign, DSAN, CAN and GVB-GD. For a fair comparison, we extract available baseline results from published papers.

***Implementation Details*** To conduct unbiased experiments, all algorithms use the same backbone network ResNet50[65] for both Office-31 dataset and Office-Home dataset. For data preprocessing, we adopt the same practice for both the first stage and the second stage. As for the first stage, data preprocessing, distribution alignment and model inference for different algorithms are all implemented according to the protocols introduced in each corresponding paper.

At the second stage, we sequentially build our proposed SPCL on top of the distribution-aligned model, *e.g.,* CAN+SPCL means that we apply SPCL to the adapted model trained with CAN algorithm. With ground truth labels for source distribution and soft pseudo-labels for target distribution, we fine-tune all convolutional and pooling layers of ImageNet[67] pretrained ResNet50 and train classifier layers from random initialization via back-propagation. Since classifier layers are trained from scratch, we set their learning rate to be ten times that of the other layers in the network. For all the domain adaptation tasks, we use minibatch Stochastic Gradient Descent (SGD) as our optimization approach. In practice, momentum is 0.9 and weight decay is $5e^{-4}$. To avoid high computational cost of grid search, we adopt the learning rate annealing strategy as in GRL[26]. Technically, the learning rate $\eta$ is adjusted by

$$\eta = \frac{\eta_0}{(1+\gamma \frac{T}{T_{max}})^{\beta}}, \tag{11}$$

where $\eta_0 = 0.01$ denotes the initial learning rate, $\gamma = 10$ and $\beta = 0.75$. We train 200 epochs at the second stage. $\frac{T}{T_{max}}$ represents the training progress increasing linearly from 0 to 1. Besides, in the implementation of $\mathcal{L}_{KL}$, the hyper-parameter *temperature* is 2. All labeled source data and unlabeled target data participate in both stages. SPCL is implemented in PyTorch and all experiments are conducted on a single GPU of NVIDIA 1080Ti with batch size of 32. For Office-31, we report results of average classification accuracy and standard error of three random trials. For Office-Home, results are obtained from the average of three runs.

## 3.2 Results on Benchmark Datasets

The classification results on Office-31 and Office-Home are respectively shown in Table 1 and Table

2.

Table1 Accuracies (%) on Office-31 for unsupervised domain adaptation methods (ResNet50)

| Method | A→W | D→W | W→D | A→D | D→A | W→A | Avg |
|---|---|---|---|---|---|---|---|
| ResNet50[65] | 68.4±0.5 | 96.7±0.5 | 99.3±0.1 | 68.9±0.2 | 62.5±0.3 | 60.7±0.3 | 76.1 |
| DDC[19] | 75.8±0.2 | 95.0±0.2 | 98.2±0.1 | 77.5±0.3 | 67.4±0.4 | 64.0±0.5 | 79.7 |
| Deep CORAL[24] | 77.7±0.3 | 97.6±0.2 | 99.7±0.1 | 81.1±0.4 | 64.6±0.3 | 64.0±0.4 | 80.8 |
| DANN[27] | 82.0±0.4 | 96.9±0.2 | 99.1±0.1 | 79.7±0.4 | 68.2±0.4 | 67.4±0.5 | 82.2 |
| ADDA[54] | 86.2±0.5 | 96.2±0.3 | 98.4±0.3 | 77.8±0.3 | 69.5±0.4 | 68.9±0.5 | 82.9 |
| JAN[48] | 85.4±0.3 | 97.4±0.2 | 99.8±0.2 | 84.7±0.3 | 68.6±0.3 | 70.0±0.4 | 84.3 |
| MADA[33] | 90.0±0.1 | 97.4±0.1 | 99.6±0.1 | 87.8±0.2 | 70.3±0.3 | 66.4±0.3 | 85.2 |
| MCD[56] | 88.6±0.2 | 98.5±0.1 | **100.0**±0.0 | 92.2±0.2 | 69.5±0.1 | 69.7±0.3 | 86.5 |
| GTA[52] | 89.5±0.5 | 97.9±0.3 | 99.8±0.4 | 87.7±0.5 | 72.8±0.3 | 71.4±0.4 | 86.5 |
| DWL[9] | 89.2 | 99.2 | **100.0** | 91.2 | 73.1 | 69.8 | 87.1 |
| iCAN[5] | 92.5 | 98.8 | **100.0** | 90.1 | 72.1 | 69.9 | 87.2 |
| CDAN+E[7] | 94.1±0.1 | 98.6±0.1 | **100.0**±0.0 | 92.9±0.2 | 71.0±0.3 | 69.3±0.3 | 87.7 |
| MDD[8] | 94.5±0.3 | 98.4±0.1 | **100.0**±0.0 | 93.5±0.2 | 74.6±0.3 | 72.2±0.1 | 88.9 |
| MetaAlign[10] | 93.0±0.5 | 98.6±0.0 | **100.0**±0.0 | 94.5±0.3 | 75.0±0.3 | 73.6±0.0 | 89.2 |
| ILA-DA[11] | 95.7 | **99.3** | **100.0** | 93.4 | 72.1 | 75.4 | 89.3 |
| E-MixNet[12] | 93.0±0.3 | 99.0±0.1 | **100.0**±0.0 | **95.6**±0.2 | **78.9**±0.5 | 74.7±0.7 | 90.2 |
| TSA[13] | **96.0** | 98.7 | **100.0** | 95.4 | 76.7 | 76.8 | 90.6 |
| DAN[18] | 83.8±0.4 | 96.8±0.2 | 99.5±0.1 | 78.4±0.2 | 66.7±0.3 | 62.7±0.2 | 81.3 |
| DAN+SPCL | 84.1±0.3 | 97.0±0.2 | 99.6±0.1 | 78.6±0.2 | 66.9±0.2 | 63.0±0.2 | 81.5 |
| DSAN[3] | 93.6±0.2 | 98.3±0.1 | **100.0**±0.0 | 90.2±0.7 | 73.5±0.5 | 74.8±0.4 | 88.4 |
| DSAN+SPCL | 94.3±0.3 | 98.5±0.1 | **100.0**±0.0 | 92.0±0.2 | 74.7±0.1 | 75.4±0.3 | 89.2 |
| GVB-GD[4] | 94.8±0.5 | 98.7±0.3 | **100.0**±0.0 | 95.0±0.4 | 73.4±0.3 | 73.7±0.4 | 89.3 |
| GVB-GD+SPCL | 95.7±0.3 | **99.3**±0.2 | **100.0**±0.0 | 95.3±0.3 | 75.0±0.3 | 75.1±0.3 | 90.1 |
| CAN[6] | 94.5±0.3 | 99.1±0.2 | 99.8±0.2 | 95.0±0.3 | 78.0±0.3 | 77.0±0.3 | 90.6 |
| CAN+SPCL | 95.5±0.1 | **99.3**±0.1 | **100.0**±0.0 | 95.5±0.1 | 78.2±0.1 | **77.6**±0.1 | **91.0** |

Table2 Accuracies (%) on Office-Home for unsupervised domain adaptation methods (ResNet50)

| Method | A→C | A→P | A→R | C→A | C→P | C→R | P→A | P→C | P→R | R→A | R→C | R→P | Avg |
|---|---|---|---|---|---|---|---|---|---|---|---|---|---|
| ResNet50[65] | 34.9 | 50.0 | 58.0 | 37.4 | 41.9 | 46.2 | 38.5 | 31.2 | 60.4 | 53.9 | 41.2 | 59.9 | 46.1 |
| DANN[27] | 45.6 | 59.3 | 70.1 | 47.0 | 58.5 | 60.9 | 46.1 | 43.7 | 68.5 | 63.2 | 51.8 | 76.8 | 57.6 |
| JAN[48] | 45.9 | 61.2 | 68.9 | 50.4 | 59.7 | 61.0 | 45.8 | 43.4 | 70.3 | 63.9 | 52.4 | 76.8 | 58.3 |
| MCD[56] | 48.9 | 68.3 | 74.6 | 61.3 | 67.6 | 68.8 | 57.0 | 47.1 | 75.1 | 69.1 | 52.2 | 79.6 | 64.1 |
| CDAN+E[7] | 50.7 | 70.6 | 76.0 | 57.6 | 70.0 | 70.0 | 57.4 | 50.9 | 77.3 | 70.9 | 56.7 | 81.6 | 65.8 |
| MDD[8] | 54.9 | 73.7 | 77.8 | 60.0 | 71.4 | 71.8 | 61.2 | 53.6 | 78.1 | 72.5 | 60.2 | 82.3 | 68.1 |
| E-MixNet[12] | 57.7 | 76.6 | 79.8 | 63.6 | 74.1 | 75.0 | 63.4 | 56.4 | 79.7 | 72.8 | 62.4 | 85.5 | 70.6 |
| TSA[13] | 57.6 | 75.8 | **80.7** | 64.3 | **76.3** | 75.1 | **66.7** | 55.7 | 81.2 | 75.7 | 61.9 | 83.8 | 71.2 |
| MetaAlign[10] | **59.3** | 76.0 | 80.2 | 65.7 | 74.7 | 75.1 | 65.7 | 56.5 | 81.6 | 74.1 | 61.1 | 85.2 | 71.3 |
| DAN[18] | 43.6 | 57.0 | 67.9 | 45.8 | 56.5 | 60.4 | 44.0 | 43.6 | 67.7 | 63.1 | 51.5 | 74.3 | 56.3 |
| DAN+SPCL | 43.9 | 57.4 | 68.5 | 46.2 | 57.2 | 60.6 | 44.3 | 44.0 | 68.2 | 64.2 | 52.0 | 74.8 | 56.8 |
| DSAN[3] | 54.4 | 70.8 | 75.4 | 60.4 | 67.8 | 68.0 | 62.6 | 55.9 | 78.5 | 73.8 | 60.6 | 83.1 | 67.6 |
| DSAN+SPCL | 56.5 | 72.3 | 76.6 | 62.5 | 70.3 | 69.3 | 64.0 | 57.1 | 79.3 | 74.2 | **62.6** | 84.0 | 69.1 |
| CAN[6] | 56.4 | 74.4 | 76.3 | 64.9 | 72.6 | 71.3 | 65.0 | 52.9 | 76.3 | 69.9 | 57.0 | 78.0 | 67.9 |
| CAN+SPCL | 57.1 | **77.0** | 78.0 | **66.9** | 73.1 | 73.8 | 65.7 | 56.4 | 78.7 | 75.2 | 59.6 | 82.0 | 70.3 |
| GVB-GD[4] | 57.0 | 74.7 | 79.8 | 64.6 | 74.1 | 74.6 | 65.2 | 55.1 | 81.0 | 74.6 | 59.7 | 84.3 | 70.4 |
| GVB-GD+SPCL | **59.3** | 76.8 | **80.7** | 66.6 | 75.9 | **75.8** | **66.7** | **57.4** | **82.9** | **76.5** | 61.8 | **85.8** | **72.2** |

SPCL is generic and it can be applied to various distribution-aligning UDA algorithms. Specifically, we adopt four UDA methods as baselines to validate the effectiveness of SPCL: DAN, DSAN, CAN and GVB-GD. Among them, DAN, DSAN and GVB-GD have been introduced in Section 1. CAN aligns both intra-class discrepancy and inter-class discrepancy with adversarial learning.

Experiment results reveal some insightful observations:

- SPCL is model-agnostic. Since the increase of combined risk $C$ during adaptation is neglected, SPCL is universally effective to improve those UDA methods of distribution alignment. On both Office-31 and Office-Home, SPCL consistently enhances classification performance of all four baselines (DAN and DSAN are statistics-based methods while CAN and GVB-GD are adversarial learning methods) in each adaptation task. Especially for challenging benchmark

Office-Home, SPCL boosts classification accuracy by large margins, *i.e.*, 1.5%, 2.4% and 1.8% on average for DSAN, CAN and GVB-GD respectively. Regardless of whether the distribution alignment is statistics-based or adversarial, SPCL decreases the upper bound of expected error $R_{D^t}$ on target domain by minimizing the combined risk.

- Note that the increase for DAN is relatively small, *i.e.*, 0.2% on Office-31 and 0.5% on Office-Home. That is because DAN has a higher inaccuracy $\rho$ of soft label. On Office-31 and Office-Home, DAN has an average classification accuracy of about 81% and 56%, while DSAN, CAN and GVB-GD achieve over 88% and 67%. Therefore, a larger $\rho$ results in a rather lower enhancement for DAN.
- SPCL is capable to outperform SOTA UDA algorithms. Prior distribution-aligning UDA methods neglect the combined risk, which increases expected error on target domain. SPCL tackles this challenge with the soft label strategy and curriculum learning strategy, optimizing the upper bound. On Office-31 and Office-Home, SPCL achieves 91.0% classification accuracy based on CAN and 72.2% accuracy with GVB-GD.

Considering that SPCL can use available UDA models that are trained and released online as $\mathcal{M}_a$ and $\mathcal{M}_a$ is fixed during the second stage training, SPCL has the same model complexity as current one-stage algorithms that utilize ResNet50 as the backbone network and the second stag training time is about 2 hours.

### 3.3 Feature Visualization

We visualize the representations of task W→A by DSAN, CAN and CAN+SPCL using t-SNE[68] in Fig.2.

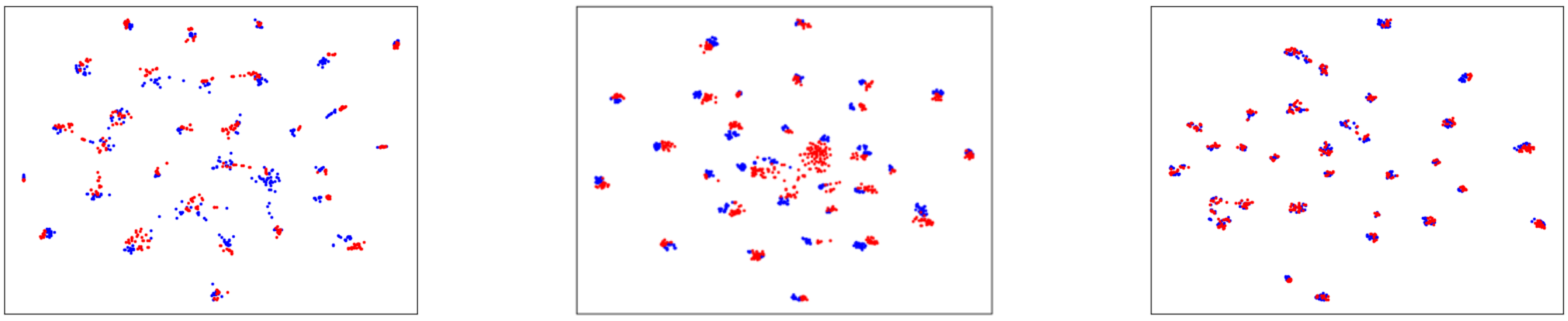

(a) DSAN (b) CAN (c) CAN+SPCL

Fig. 2 Visualization with t-SNE on Office-31 task W→A (blue: W, red: A)

CAN aligns the source and target domain better than DSAN with higher intra-class compactness but target categories are not discriminated well. With SPCL, target representations with different categories are more dispersed, demonstrating the benefit of soft pseudo-label strategy and curriculum learning (adaptive weighting) strategy.

Overall, the aforementioned experiment results validate the effectiveness of our proposed SPCL and its superior performance.

## 4 Ablation Study

### 4.1 Soft Label Strategy

To reduce inaccuracy pseudo-label ratio $\rho$, we adopt soft pseudo-labels instead of one-hot predictions from the adapted model. As analyzed in Section 2.3, we are motivated by two insights. Firstly, model predictions during training iterations are inaccurate and inconsistent. Secondly, soft labels can regularize label smoothing and convey task-related semantic information.

To prove the higher label accuracy of adapted model, we visualize the CAN model performance during the training iterations at the first stage on task C→A, P→C and R→P.

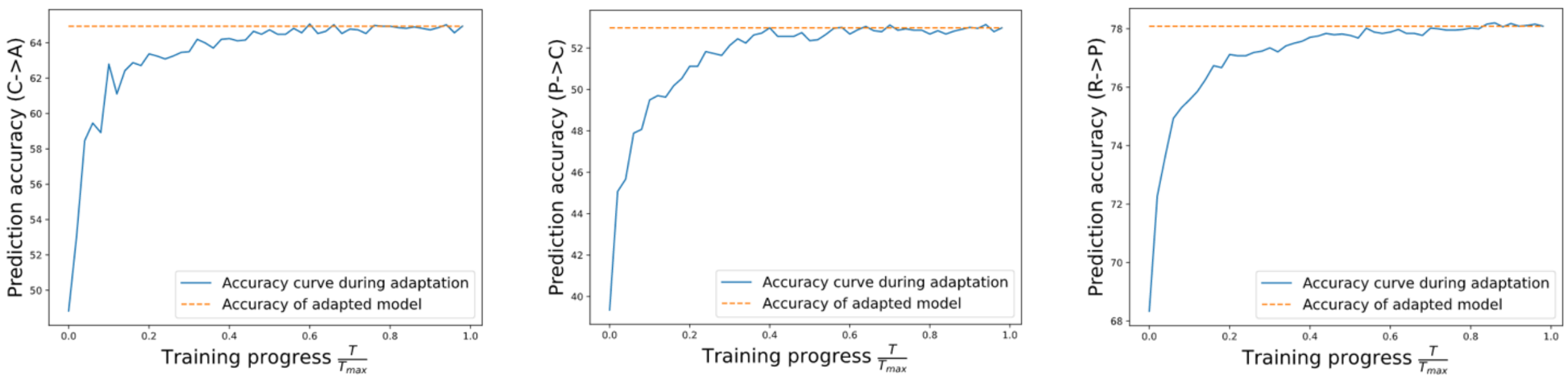

(a) (b) (c)

Fig. 3 Prediction accuracy of CAN on target domain at the first stage of Office-Home adaptation task C→A, P→C and R→P

As shown in Fig.3, the classification performance is improved in each task as the training proceeds at the first stage. In the end, the adapted model can provide pseudo-labels with rather high confidence and accuracy.

Then we investigate different label strategies (soft pseudo-label and hard pseudo-label) for target domain with CAN+SPCL and GVB-GD+SPCL on both benchmark datasets Office-31 and Office-Home.

Results are shown in Table 3, which demonstrates our insights.

Table3 Ablation study of different label strategies for target domain on Office-31 and Office-Home

| Dataset | Method | Label strategy | Average accuracy |
|---|---|---|---|
| Office-31 | CAN(Baseline) | / | 90.6 |
| | CAN+SPCL | Hard pseudo-label | 90.4 |
| | | Soft pseudo-label model | 91.0 |
| | GVB-GD(Baseline) | / | 89.3 |
| | GVB-GD+SPCL | Hard pseudo-label | 88.9 |
| | | Soft pseudo-label model | 90.1 |
| Office-Home | CAN(Baseline) | / | 67.9 |
| | CAN+SPCL | Hard pseudo-label | 67.5 |
| | | Soft pseudo-label model | 70.3 |
| | GVB-GD(Baseline) | / | 70.4 |
| | GVB-GD+SPCL | Hard pseudo-label | 69.9 |
| | | Soft pseudo-label model | 72.2 |

Training the second-stage network with hard pseudo-label results in inferior classification accuracy compared with baseline model *i.e.*, CAN and GVB-GD. As hard pseudo-label increases inaccuracy $\rho$, paying more attention to one-hot labels only forces the network to learn inaccurate distribution. Instead, utilizing soft labels reduces network performance on target domain by transferring task-related semantic information.

## 4.2 Curriculum Learning Strategy

In order to minimize combined risk $C_l$, we shift the training focus from source distribution to target distribution by adaptively adjusting the loss weighting $\lambda$ within the range of $[0,1]$. As analyzed in Section 2.4, there are two advantages. Firstly, learning source distribution at early stages guides the training network towards a solid convergence. Secondly, focusing on target distribution avoids overfitting on source domain and improves classification accuracy on target domain. In Fig.4 we validate the effectiveness of SPCL in solving the overfitting problem.

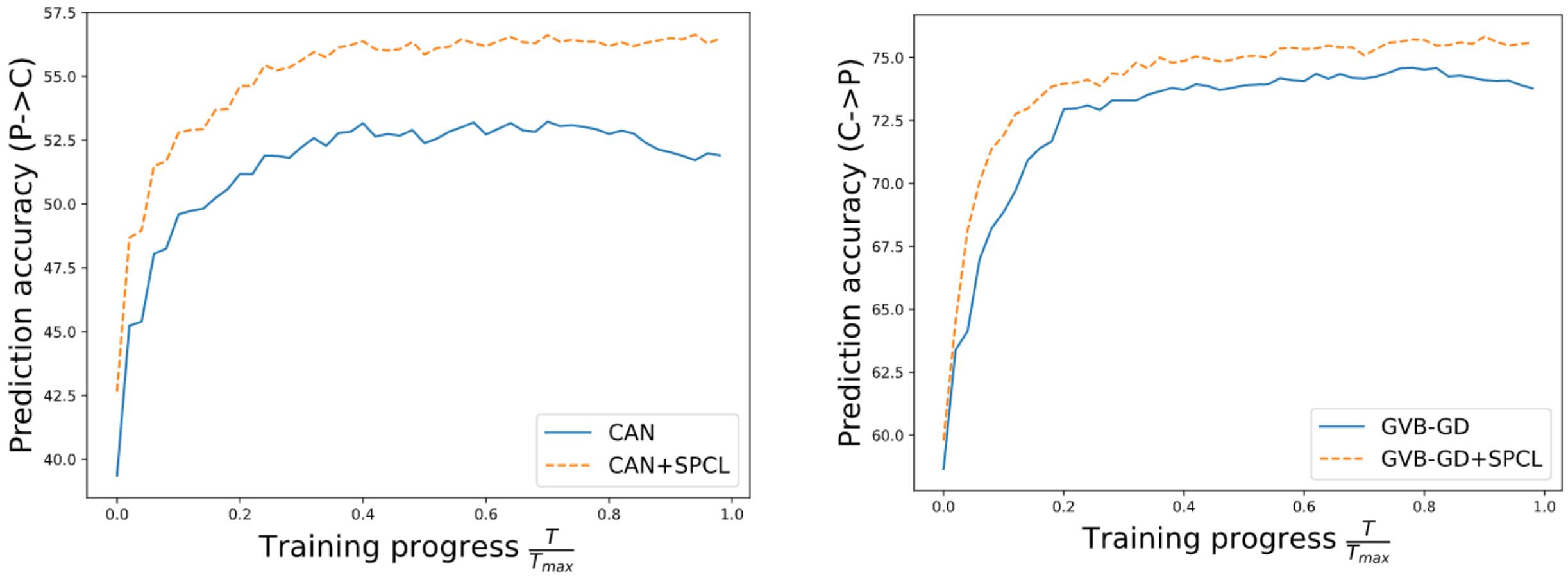

(a) (b)

Fig. 4 Prediction accuracy on target domain of Office-Home adaptation task P→C and C→P

To facilitate the understanding of our curriculum learning strategy, we explore several adjusting mechanisms for $\lambda$, *i.e.*, progress-related and progress-unrelated ones. We apply the weighting mechanisms to the model adapted by CAN and conduct experiments for three runs on Office-31. Results are shown in Table 4. We denote training progress by $r = \frac{T}{T_{max}}$, which increases linearly from 0 to 1.

Table4 Ablation study of different weighting strategies of CAN+SPCL on Office-31

| Strategy type | Mechanism | $\lambda$ | Average accuracy |
|---|---|---|---|
| Decrement | Steep exponential decrement | $2-\frac{2}{1+e^{-10r}}$ | 87.1 |
| Fixed | Source-dominant weighting | 0.2 | 87.7 |
| | Equal weighting | 0.5 | 88.1 |
| | Target-dominant weighting | 0.8 | 88.2 |
| Increment | Flat exponential increment | $e^{\ln(2)r^3}-1$ | 88.6 |
| | Cosine increment | $1-\cos(\frac{\pi}{2}r)$ | 89.3 |
| | Linear increment | $r$ | 89.8 |
| | Steep exponential increment (Ours) | $\frac{2}{1+e^{-10r}}-1$ | **91.0** |

We present progress-unrelated mechanisms, *i.e.*, fixed $\lambda$ value (source-dominant weighting, equal weighting for source and target distribution and target-dominant weighting). Besides, progress-related mechanisms are implemented by adjusting $\lambda$ exponentially, linearly, *etc*. As shown in Table 4, the increment strategies (cosine increment, linear increment and exponential increment) perform better than both fixed (source-dominant weighting, equal weighting and target-dominant weighting) and decrement strategy. It proves our intuition that shifting training focus to target domain improves classification accuracy. Decrement strategy gains the worst performance, demonstrating the negative effect of overfitting to the source distribution. Fixed strategies validate the benefit of focusing on target distribution learning. Among the increment strategies, our proposed steep exponential increment strategy has the highest accuracy.

We visualize how $\lambda$ varies with the progress of training in Fig.5.

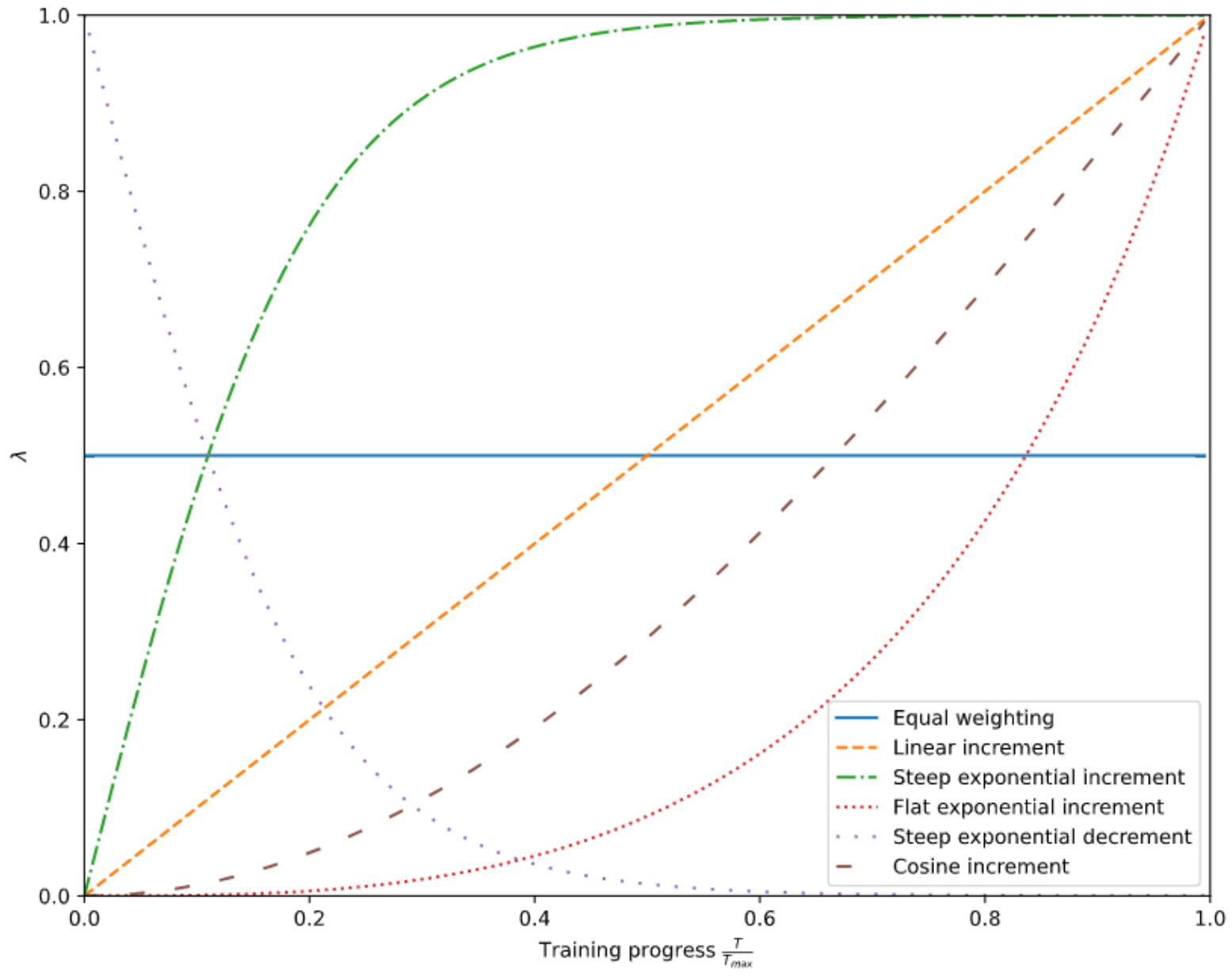

Fig. 5. Visualization of different mechanism

Notably, training focuses on target domain when $\lambda > 0.5$. Our proposed strategy with exponential increment prolongs the epochs of target distribution learning. That's the reason why it gains the highest accuracy.

## 4.3 Network Initialization

Instead of continuing the second stage training with the distribution-aligned model, we use an ImageNet pretrained ResNet50 for better model generalization ability. During the alignment (both explicit statistics-based method and adversarial learning algorithm), domain-specific characteristics and domain-invariant representations are assumed to be orthogonal and the domain-specific part is diminished, which contains task-related information though[69]. Therefore, training adapted model with SPCL will gain limited improvement. Besides, being trapped in a saddle point or local optimum disables the model from further performance improvement[70-71]. To validate that insight, we initialize the second stage network with adapted model weight or ImageNet pretrained model weight and conduct experiments on both Office-31 and Office-Home. Results are shown in Table 5.

Table5 Ablation study of different network initialization on Office-31 and Office-Home

<table>
<tr><th>Dataset</th><th>Method</th><th>Label strategy</th><th>Average accuracy</th></tr>
<tr><td rowspan="6">Office-31</td><td>CAN(Baseline)</td><td>ImageNet pretrained model</td><td>90.6</td></tr>
<tr><td rowspan="2">CAN+SPCL</td><td>Adapted model from the first stage</td><td>90.7</td></tr>
<tr><td>ImageNet pretrained model</td><td>91.0</td></tr>
<tr><td>GVB-GD(Baseline)</td><td>ImageNet pretrained model</td><td>89.3</td></tr>
<tr><td rowspan="2">GVB-GD+SPCL</td><td>Adapted model from the first stage</td><td>89.5</td></tr>
<tr><td>ImageNet pretrained model</td><td>90.1</td></tr>
<tr><td rowspan="6">Office-Home</td><td>CAN(Baseline)</td><td>ImageNet pretrained model</td><td>67.9</td></tr>
<tr><td rowspan="2">CAN+SPCL</td><td>Adapted model from the first stage</td><td>68.2</td></tr>
<tr><td>ImageNet pretrained model</td><td>70.3</td></tr>
<tr><td>GVB-GD(Baseline)</td><td>ImageNet pretrained model</td><td>70.4</td></tr>
<tr><td rowspan="2">GVB-GD+SPCL</td><td>Adapted model from the first stage</td><td>70.9</td></tr>
<tr><td>ImageNet pretrained model</td><td>72.2</td></tr>
</table>

Classification performance is inferior if the network continues training with distribution-aligned model instead of with ImageNet pretrained model. That demonstrates our motivation that initializing the second stage training network with ImageNet pretrained model parameters offers a start point with better generalization ability.

## 5 Conclusion

In this paper, we minimize the combined risk for unsupervised domain adaptation. As compared with previous researches, which enhance distribution alignment with model predictions on target domain by either explicit statistics-based method or adversarial learning algorithm, our approach focuses on two unexplored challenges, *i.e.*, inaccurate category predictions on target domain and overfitting of the source distribution. Through theoretical analysis, the combined risk $C$ on source domain and target domain can be estimated by inaccuracy ratio $\rho$ of pseudo-labels, and the summation of combined risk $C_l$ over source domain and pseudo-labeled target domain. Accordingly, we propose a model-agnostic two-stage framework SPCL to reduce those two terms, thus successfully lowering the upper bound of expected errors on target domain. In order to minimize $\rho$, we design a Soft Pseudo-label strategy, using soft labels with a rather high confidence provided by the aligned model at the cost of a little higher (about 65% of the one-stage model) memory usage. To decrease $C_l$, we adopt a Curriculum Learning strategy to gradually shift training focus from source distribution to target distribution. Extensive experiments are conducted on two standard benchmarks and validate that SPCL is universally effective for various distribution-aligning UDA algorithms with boosted classification accuracy.